\documentclass[10pt,twocolumn,letterpaper]{article}

\usepackage{cvpr}              

\usepackage[dvipsnames]{xcolor}

\definecolor{cvprblue}{rgb}{0.21,0.49,0.74}
\usepackage[pagebackref,breaklinks,colorlinks,citecolor=cvprblue]{hyperref}
\usepackage{multirow}

\def\paperID{11051} 
\def\confName{CVPR}
\def\confYear{2024}

\makeatletter
\def\blfootnote{\gdef\@thefnmark{}\@footnotetext}
\makeatother

\title{DEADiff: An Efficient Stylization Diffusion Model with Disentangled Representations\vspace{-5mm}}

\author{
Tianhao Qi\textsuperscript{\rm 1,2} \quad Shancheng Fang\textsuperscript{\rm 1} \quad Yanze Wu\textsuperscript{\rm 2} \quad Hongtao Xie\textsuperscript{\rm 1*} \quad Jiawei Liu\textsuperscript{\rm 2} \quad Lang Chen\textsuperscript{\rm 2} \\
Qian He\textsuperscript{\rm 2} \quad Yongdong Zhang\textsuperscript{\rm 1}\\
\textsuperscript{\rm 1}University of Science and Technology of China \quad \textsuperscript{\rm 2}ByteDance Inc.\\
{\tt\small qth@mail.ustc.edu.cn \quad \{fangsc, htxie, zyd73\}@ustc.edu.cn}\\
{\tt\small \{wuyanze.cs, liujiawei.cc22, chenlang.cl, heqian\}@bytedance.com}
}

\begin{document}
\twocolumn[{%
\renewcommand\twocolumn[1][]{#1}%
\maketitle
\vspace{-12mm}
\begin{center}
    \centering
    \includegraphics[width=\linewidth]{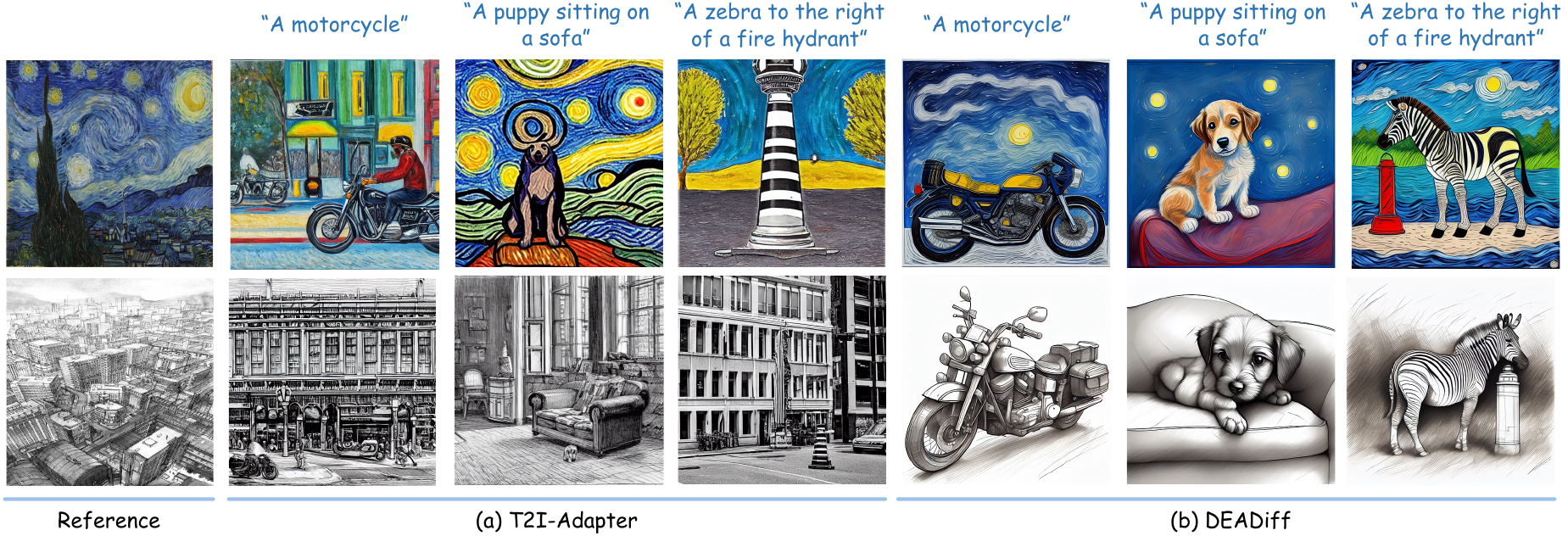}
    \vspace{-7mm}
    \captionof{figure}{Given a style reference image, \textbf{DEADiff} is capable of synthesizing new images that resemble the style and are faithful to text prompts simultaneously. However, previous encoder-based methods (\textit{i.e.}, T2I-Adapter~\cite{mou2023t2i}) significantly impair the text controllability of the diffusion-based text-to-image models.}
    \label{fig:teaser}
\end{center}
}]
\begin{abstract}
\vspace{-3mm}
The diffusion-based text-to-image model harbors immense potential in transferring reference style. However, current encoder-based approaches significantly impair the text controllability of text-to-image models while transferring styles. In this paper, we introduce \textit{DEADiff} to address this issue using the following two strategies: 1) a mechanism to decouple the style and semantics of reference images. The decoupled feature representations are first extracted by Q-Formers which are instructed by different text descriptions. Then they are injected into mutually exclusive subsets of cross-attention layers for better disentanglement. 2) A non-reconstructive learning method. The Q-Formers are trained using paired images rather than the identical target, in which the reference image and the ground-truth image are with the same style or semantics. We show that DEADiff attains the best visual stylization results and optimal balance between the text controllability inherent in the text-to-image model and style similarity to the reference image, as demonstrated both quantitatively and qualitatively. Our project page is~\href{https://tianhao-qi.github.io/DEADiff/}{https://tianhao-qi.github.io/DEADiff/}.
\blfootnote{\hspace{-2em}*Corresponding authors.}

\end{abstract}
\vspace{-4mm}
\section{Introduction}
\label{sec:intro}

Recently, Diffusion models~\cite{ramesh2022hierarchical,saharia2022photorealistic,rombach2022high} in text-to-image generation have sparked widespread research due to their astounding performance. As Diffusion models are notoriously known for lacking enhanced controllability, how to stably and reliably guide them to adhere to a predetermined style defined by a reference image becomes intractable.

Taking into account both effectiveness and efficiency, a prevalent method for style transferring is the approach centered around an additional encoder~\cite{huang2023composer,mou2023t2i,zhao2023uni,li2023blip-diffusion,ye2023ip,wang2023styleadapter}. The encoder-based methods typically train an encoder to encode a reference image to informative features, which are then injected into the Diffusion model as its guided condition. Note that the encoder-based methods are quite efficient due to a single-pass computation, compared with the optimization-based methods that require multiple-iteration learning~\cite{gal2022image,zhang2023inversion,ruiz2023dreambooth,kumari2023multi,hu2021lora,sohn2023styledrop}. Through such an encoder, highly abstract features can be extracted to effectively describe the style of the reference image. These rich style features enable the Diffusion model to accurately understand the style of the reference image it needs to synthesize, as shown on the left side of~\cref{fig:teaser} where a typical method~(T2I-Adapter~\cite{mou2023t2i}) can generate naturally faithful reference styles. However, this approach also introduces a particularly vexing issue: while it allows the model to follow the style of the reference image, it significantly diminishes the model's performance in understanding the semantic context of text conditions.

The loss of text controllability primarily stems from two aspects. On the one hand, the encoder extracts information that couples style with semantics, rather than purely style features. Specifically, previous methods lack an effective mechanism in their encoders to distinguish between image style and image semantics. Therefore the extracted image features inevitably encompass both stylistic and semantic information. This image semantics conflicts with the semantics in the text conditions, leading to a weakened control over text-based conditions. On the other hand, previous methods treat the learning process of the encoder as a reconstruction task, where the ground-truth of the reference image is the image itself. Compared to training a text-to-image model to follow text descriptions, learning from the reconstruction of reference images is typically easier. Consequently, under the reconstruction task, the model tends to focus on the reference image, while neglecting the original text condition in the text-to-image model.


Concerning the above problems, we thus propose \textit{DEADiff} to efficiently transfer reference style to synthetic images without the loss of controllability of text condition. The \textit{DEADiff} consists of two components. Firstly, we decouple the style from the semantics in the reference image from the aspects of feature extraction and feature injection. For feature extraction, a dual decoupling representation extraction mechanism (DDRE) is proposed that utilizes Q-Former~\cite{li2023blip} to obtain style and semantic representations from the reference image. The Q-Former is instructed by ``style" and ``content" conditions to selectively extract features that align with the given instructions. For feature injection, we introduce a disentangled conditioning mechanism to inject decoupled representations into mutually exclusive subsets of cross-attention layers for better disentanglement, which is inspired by that different cross-attention layers in the Diffusion U-Net express distinct responses to style and semantics, as demonstrated in~\cite{voynov2023p+}. Secondly, we propose a non-reconstruction training paradigm that learns from paired synthetic images. Specifically, the Q-Former instructed by the ``style" condition is trained using paired images with the same style as the reference image and the ground-truth image, respectively. Meanwhile, the Q-Former instructed by the ``content" condition is trained by images with the same semantics but different styles. 

With the style and semantics decoupling mechanism and the non-reconstruction training objective, our \textit{DEADiff} can successfully imitate the style of the reference image, and be faithful to various text prompts, as illustrated in~\cref{fig:teaser}~(b). Compared with the optimization-based methods, our method is more efficient while simultaneously maintaining exceptional style transfer capabilities. In contrast to traditional encoder-based methods, our approach can effectively preserve text control ability. Besides, \textit{DEADiff} eliminates the need for manually adjusting trivial parameters to obtain satisfactory styles, something like feature fusion weight that is typically required by previous methods (\eg, T2I-Adapter).

In summary, our contributions are threefold:

\begin{itemize}
    \item We propose a dual decoupling representation extraction mechanism to separately obtain style and semantic representations of the reference image, alleviating the problem of semantics conflict between text and reference images from the perspective of learning tasks.
    \item We introduce a disentangled conditioning mechanism that allows different parts of the cross-attention layers to be responsible for the injection of image style/semantic representation separately, reducing the semantics conflict further from the perspective of model structure.
    \item We build two paired datasets to aid the DDRE mechanism using the non-reconstruction training paradigm.
\end{itemize}
\section{Related Work}
\label{sec:related}

\subsection{Diffusion-based Text-to-Image Generation}
In recent years, diffusion models have achieved great success in image generation. 
Diffusion Probabilistic Models (DPMs) \cite{sohl2015deep} are proposed to learn to restore the target data distributions destroyed by the forward diffusion process.
DPMs have attracted increasing attention in the community of image synthesis since the initial diffusion-based image generation works \cite{ho2020denoising,dhariwal2021diffusion,song2020denoising} prove their powerful generation capacity. 
Latest diffusion models \cite{ramesh2022hierarchical,saharia2022photorealistic,rombach2022high} further achieve state-of-the-art performance on text-to-image generation, which benefits from large-scale pre-training.
These methods use U-Net~\cite{ronneberger2015u} as the diffusion model, in which cross-attention layers are utilized for injecting the text features extracted from the pre-trained encoders \cite{radford2021learning, t5}.
Especially, Latent Diffusion Models (LDMs) \cite{rombach2022high}, which are also known as Stable Diffusion (SD) models, transfer the diffusion process to a low-resolution latent space through a pre-trained auto-encoder and achieve efficient high-resolution text-to-image generation. 
Considering the great success of diffusion-based text-to-image (T2I) generation models, abundant of recent diffusion methods \cite{zhang2023adding,ye2023ip,li2023blip-diffusion} focus on using more conditions from a reference image. 
One typical representative is the style, which is the main concern of this paper.

\subsection{Stylized Image Generation with T2I Models}
Stylized image generation has widely studied based on pre-trained deep convolutional or transformer-based neural networks \cite{an2021artflow,chen2021artistic,deng2022stytr2,gatys2016image,kolkin2019style, park2019arbitrary, wu2021styleformer}, which have made substantial advancements, leading to numerous practical applications.

Witnessed by the power of large-scale Text-to-image models, how to utilize these models to fulfill stylized image generation with better quality and more flexibility is an exciting topic to explore. Textual inversion-based methods~\cite{gal2022image,zhang2023inversion} project the style image into a learnable embedding of the text token space.
Unfortunately, the problem of information loss, stemming from the mapping from visual to text modalities, presents a significant challenge to the learned embedding in accurately rendering the style of the reference image with user-defined prompts.
In contrast, DreamBooth~\cite{ruiz2023dreambooth} and Custom Diffusion~\cite{kumari2023multi} can synthesize images that better capture the style of the reference image by optimizing all or partial parameters of the diffusion model.
Nevertheless, the cost is the decreased fidelity to text prompts resulting from the severe overfitting.
Currently, parameter-efficient fine-tuning provides a more effective approach for stylized image generation without impacting the diffusion model's fidelity to text prompts, such as InST~\cite{zhang2023inversion}, LoRA~\cite{hu2021lora} and StyleDrop~\cite{sohn2023styledrop}.
However, while these optimization-based methods can customize styles, they all require minutes to hours to fine-tune the model for each input reference image.
The additional computational and storage overhead impedes the practicality of these methods in real-world production.

Thus, some optimization-free methods \cite{huang2023composer,wang2023styleadapter,mou2023t2i} are proposed to extract style features from the reference image through designed image encoders.
Among them, T2I-Adapter-Style \cite{mou2023t2i} and IP-Adapter \cite{ye2023ip} 
use Transformer \cite{vaswani2017attention} as the image encoder with CLIP \cite{radford2021learning} image embeddings as input, and utilize the extracted image features through U-Net cross-attention layers. 
BLIP-Diffusion \cite{li2023blip-diffusion} builds a Q-Former \cite{li2023blip} to transform the image embeddings to text embedding space and input them to the text encoder of the diffusion model.
Those methods use whole image reconstruction \cite{mou2023t2i,ye2023ip} or object reconstruction \cite{li2023blip-diffusion} as the training objective, resulting in both the content and style information being extracted from the reference image.
To make the image encoders focus on extracting style features, StyleAdapter \cite{wang2023styleadapter} and ControlNet-shuffle \cite{zhang2023adding} shuffle the patch or pixel of the reference image and could generate various content with the target style.
\section{Method}

\begin{figure}
    \centering
    \includegraphics[width=1.0\linewidth]{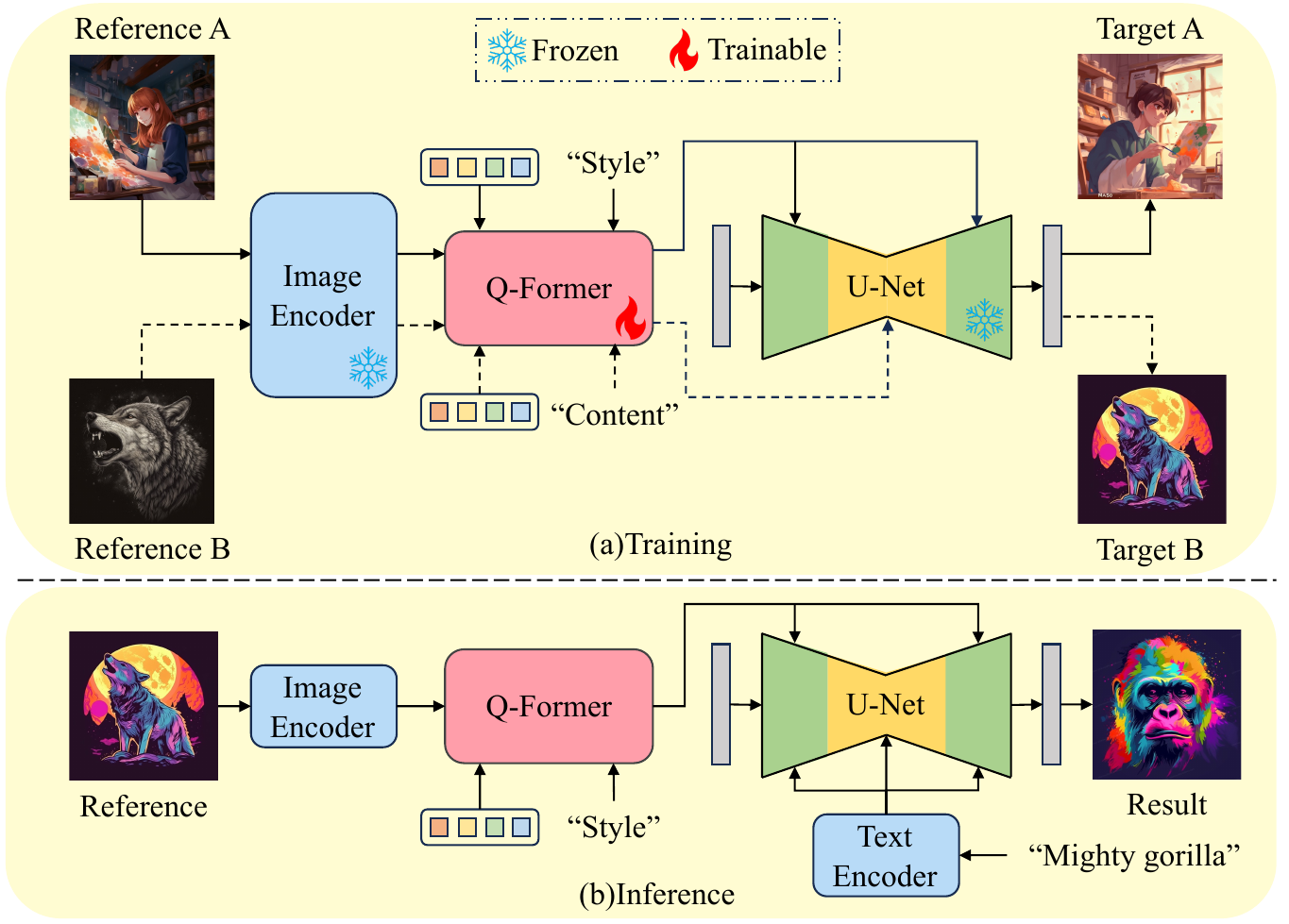}
    \caption{The training and inference paradigm of \textit{DEADiff}. We use proprietary paired datasets for training Q-Former to extract disentangled representations under conditions ``style" and ``content", which are injected into mutually exclusive cross-attention layers.}
    \label{fig:pipeline}
    \vspace{-5mm}
\end{figure}

\subsection{Preliminary}
SD is a type of latent diffusion model~\cite{rombach2022high}, which performs a sequence of gradual denoising operations within the latent space and remaps the denoised latent code into the pixel space, thereby generating the final output image.
During the training process, SD initially casts an input image $x$ into a latent code $z$ via a Variational Auto-Encoder~\cite{kingma2013auto}.
In subsequent stages, the noised latent $z_t$ at timestep $t$ serves as the input for the denoising U-Net $\epsilon_{\theta}$, which undertakes interaction with text prompts $c$ via cross-attention.
The supervision for this process is ensured by the following objective:
\begin{equation}
    L = \mathbb{E}_{z, c, \epsilon \sim \mathcal{N}(0,1), t}\left[\left\|\epsilon-\epsilon_{\theta}\left(z_{t}, t, c\right)\right\|_{2}^{2}\right],
    \label{equ:1}
\end{equation}
where $\epsilon$ represents a random noise sampled from the standard Gaussian distribution.

\subsection{Dual Decoupling Representation Extraction}
\label{sec:3.2}

Taking inspiration from BLIP-Diffusion~\cite{li2023blip-diffusion}, which learns the subject representations through synthetic image pairs with different background to avoid trivial solution, we integrate two auxiliary tasks that utilize Q-Formers as representation filters nesting within a non-reconstructive paradigm.
This enables us to implicitly discern disentangled representations of both style and content within an image.

On the one hand, we sample a pair of distinct images, both maintaining the same style but serving as the reference and target respectively for the Stable Diffusion (SD) generation process, as depicted in pair A of~\cref{fig:pipeline}(a).
The reference image is fed into the CLIP image encoder, whose output interacts with the learnable query tokens of the Q-Former~\cite{li2023blip} and its input text through cross-attention.
For this process, we settle on the word ``style" as the input text in anticipation of generating text-aligned image features as output.
This output, which encapsulates the style information, is then coupled with the caption detailing the content of the target image and provided for conditioning to the denoising U-Net.
The impetus for this prompt composition strategy aims to better disentangle the style from the content caption allowing the Q-Former to focus more on the extraction of style-centric representations.
This learning task is defined as the style representation extraction, abbreviated as STRE.

On the other hand, we incorporate a corresponding and symmetric content representation extraction task, referred to as SERE.
As shown in pair B of~\cref{fig:pipeline}(a), we select two images that share the same subject matter but exhibit distinct styles, which are assigned as the reference and target images.
Importantly, we replace the input text of the Q-Former with the word ``content" to extract associated content-specific representations.
To acquire unadulterated content representations, we supply the output of the query token by the Q-Former and the text style words of the target image, concurrently, as the conditioning for the denoising U-Net.
In this approach, the Q-Former will sieve out content unrelated information nested within the CLIP image embeddings while generating the target image.

Simultaneously, we incorporate a reconstruction task into the entire pipeline.
The conditioning prompt consists of the query tokens processed by the ``style" Q-Former and ``content" Q-Former for this learning task.
In this way, we can ensure that Q-Formers do not neglect essential image information, considering the complementary relationship between content and style.

\subsection{Disentangled Conditioning Mechanism}

\begin{figure}[t]
    \centering
    \includegraphics[width=1.0\linewidth]{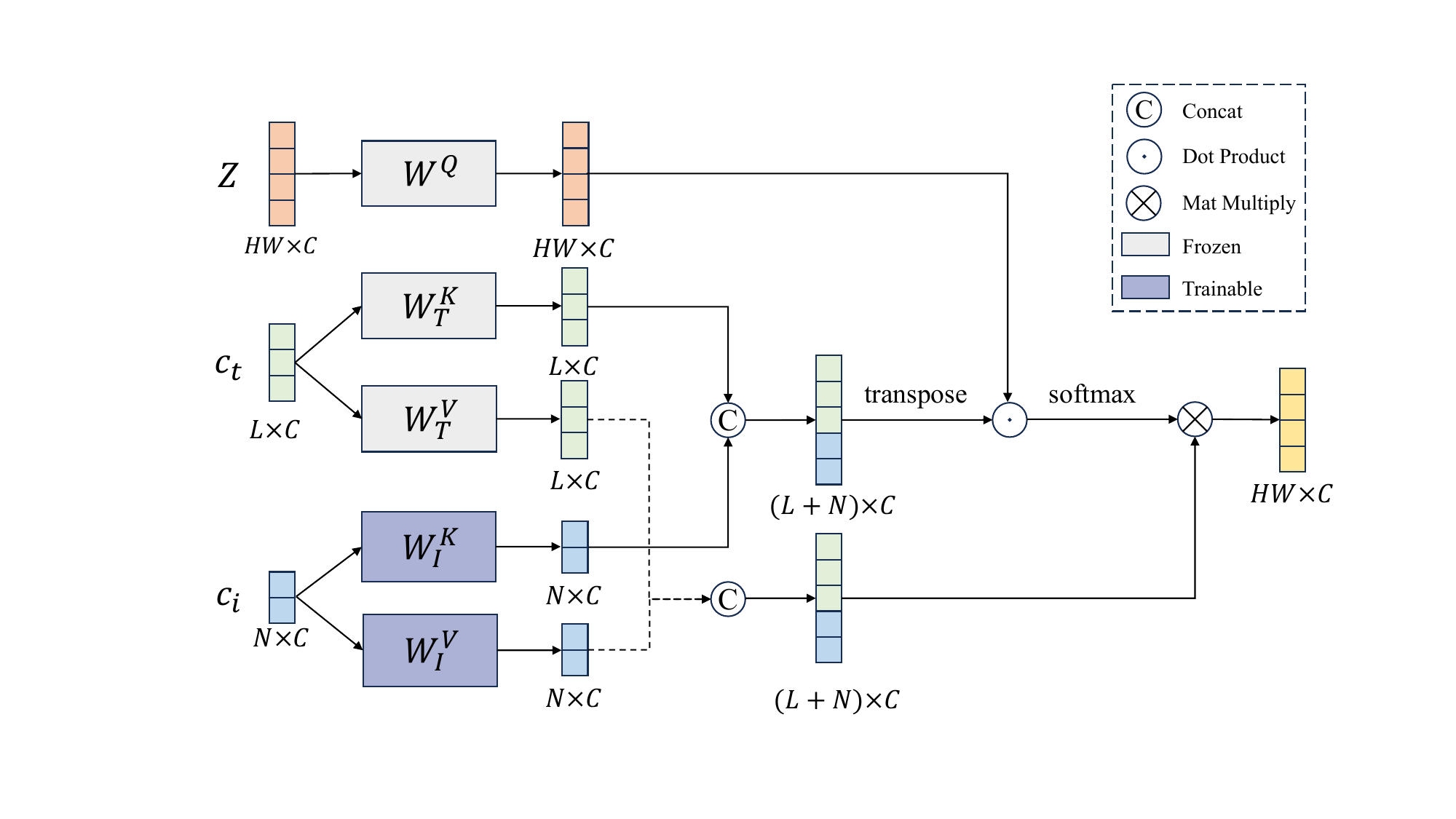}
    \caption{The illustration of our proposed joint text-image cross-attention layer.}
    \label{fig:cross-attn}
    \vspace{-5mm}
\end{figure}

Motivated by the observation in~\cite{voynov2023p+} that different cross-attention layers in the denoising U-Net dominate different attributes of the synthesized image, we introduce an innovative Disentangled Conditioning Mechanism (DCM).
In essence, DCM adopts a strategy that conditions the coarse layers with lower spatial resolution on semantics, while the fine layers with higher spatial resolution are conditioned on the style.
As illustrated in~\cref{fig:pipeline}(a), we only inject the output queries of the Q-Former with ``style" conditions to fine layers, which respond to local area features rather than global semantics.
This structural adaptation propels the Q-Former to extract more style-oriented features, such as strokes, textures, and colors of the image when inputted with “style” conditions, while diminishing its focus on global semantics.
This strategy hence enables a more effective decoupling of style and semantic features.
Simultaneously, to make the denoising U-Net support image features as conditions, we devise a joint text-image cross-attention layer, as demonstrated in~\cref{fig:cross-attn}.
In a manner akin to IP-Adapter~\cite{ye2023ip}, we include two trainable linear projection layers $W_{I}^{K}$, $W_{I}^{V}$ to process image features $c_i$, in conjunction with frozen ones $W_{T}^{K}$, $W_{T}^{V}$ for text features $c_t$.
However, instead of executing cross-attention for image and text features independently, we concatenate the key and value matrices from text and image features respectively, subsequently initiating a single cross-attention operation with U-Net query features $Z$.
Formally, the formulation of this combined text-image cross-attention process can be expressed as follows:
\begin{align}
    Q&=ZW^{Q}, \\
    K&=Concat(c_{t}W_{T}^{K}, c_{i}W_{I}^{K}), \\
    V&=Concat(c_{t}W_{T}^{V}, c_{i}W_{I}^{V}), \\
    Z^{new}&=Softmax(\frac{QK^{T}}{\sqrt{d}})V.
\end{align}

\subsection{Paired Datasets Construction}
\label{sec:3.3}
Preparing a pair of images with the same style or subject as stated in~\cref{sec:3.2} is a non-trivial work.
Fortunately, existing state-of-the-art text-to-image models have demonstrated a strong fidelity to given text prompts.
Therefore, we manually create a list of text prompts by combining subject words and style words, and utilize a pre-trained model to construct two paired image datasets - one with samples of the same style and the other with samples of the same subject.
Formally, the construction of the paired datasets involves the following three steps:

\noindent\textbf{Step 1: Text prompt combination.} We have listed nearly 12,000 subject words that span across four major categories: characters, animals, objects, and scenes.
Additionally, we have noted nearly 700 style words that include attributes such as artistic styles, artists, brushstrokes, shadows, shots, resolutions, and visual angles.
Then, every subject word is assigned approximately 14 style words on average from all style words, and the combination forms the final text prompts used for the text-to-image model.

\noindent\textbf{Step 2: Image generation and collection.}
After combining text prompts with subject words and style words, we have obtained over 160 thousand prompts.
Subsequently, all the text prompts are sent to Midjourney, a leading text-to-image generation product, to synthesize corresponding images.
As a characteristic of Midjourney, the direct output of a given prompt embraces 4 images with resolution $512\times 512$.
We upsample each image to resolution $1024\times 1024$ and store it with the given prompt.
Due to redundancy in data collection, we ultimately collected a total of 1.06 million image-text pairs.

\noindent\textbf{Step 3: Paired images selection.} We observe that even with the same style words, there are significant differences in images generated with different subject words.
In light of this, for the style representation learning task, we use two distinct images synthesized with the same prompt, which serve as the reference and target respectively, as illustrated in Figure ~\cref{fig:pipeline}(a).
To achieve this goal, we store images with the same prompt as a single item and randomly select two images during each iteration.
In terms of the content representation learning task depicted in~\cref{fig:pipeline}(b), we pair images with the same subject word but different style words as a single item.
Ultimately, we have obtained one dataset with over 160000 items for the former task and another one with 1.06 million items for the latter task.

\subsection{Training and Inference.}

We employ the loss function depicted in~\cref{equ:1} to supervise the aforementioned three learning tasks.
During the training process, only the Q-Former and the newly added linear projection layers are optimized.
The inference process is illustrated as shown in~\cref{fig:pipeline}(b).

\section{Experiment}

\subsection{Experiment Settings}
\textbf{Implementation Details.} We adopt Stable Diffusion v1.5 as our base text-to-image model, which comprises a total of 16 cross-attention layers.
We number them from 0 to 15 in the order from input to output and define layers 4-8 as coarse layers that are used for injecting image content representation.
Accordingly, the other layers are defined as fine layers used for injecting image style representation.
We utilize ViT-L/14 from CLIP~\cite{radford2021learning} as the image encoder and keep the number of learnable query tokens of the Q-Former consistent with BLIP-Diffusion, \textit{i.e.}, 16.
We adopt two Q-Formers to separately extract semantic and style representations, to encourage them to focus on their own tasks.
For the sake of fast convergence, we initialize the Q-Former with the pre-trained model provided by BLIP-Diffusion~\cite{li2023blip-diffusion} in HuggingFace\footnote{\href{https://huggingface.co/salesforce/blipdiffusion}{https://huggingface.co/salesforce/blipdiffusion}}.
In terms of the additional projection layers $W_{I}^{K}$, $W_{I}^{V}$, we initialize them with the parameters of $W_{T}^{K}$, $W_{T}^{V}$.
During training, we set the sampling ratio of the three learning tasks as stated in~\cref{sec:3.2} to 1:1:1, to train the style Q-Former and content Q-Former equally.
We fix the parameters of the image encoder, text encoder, and original U-Net\cite{ronneberger2015u}, and only update the parameters of the Q-Former, 16 learnable queries, and the additional projection layers $W_{I}^{K}$, $W_{I}^{V}$.
The models are trained with a total batch size of 512 on 16 A100-80G GPUs.
We employ AdamW~\cite{loshchilov2018decoupled} as the optimizer with a learning rate of $1e-4$ and train for 100000 iterations.
As for inference, we adopt the DDIM~\cite{song2020denoising} sampler with 50 steps.
The guidance scale for classifier-free guidance~\cite{ho2022classifier} is 8.


\begin{figure*}
    \centering
    \includegraphics[width=1.0\linewidth]{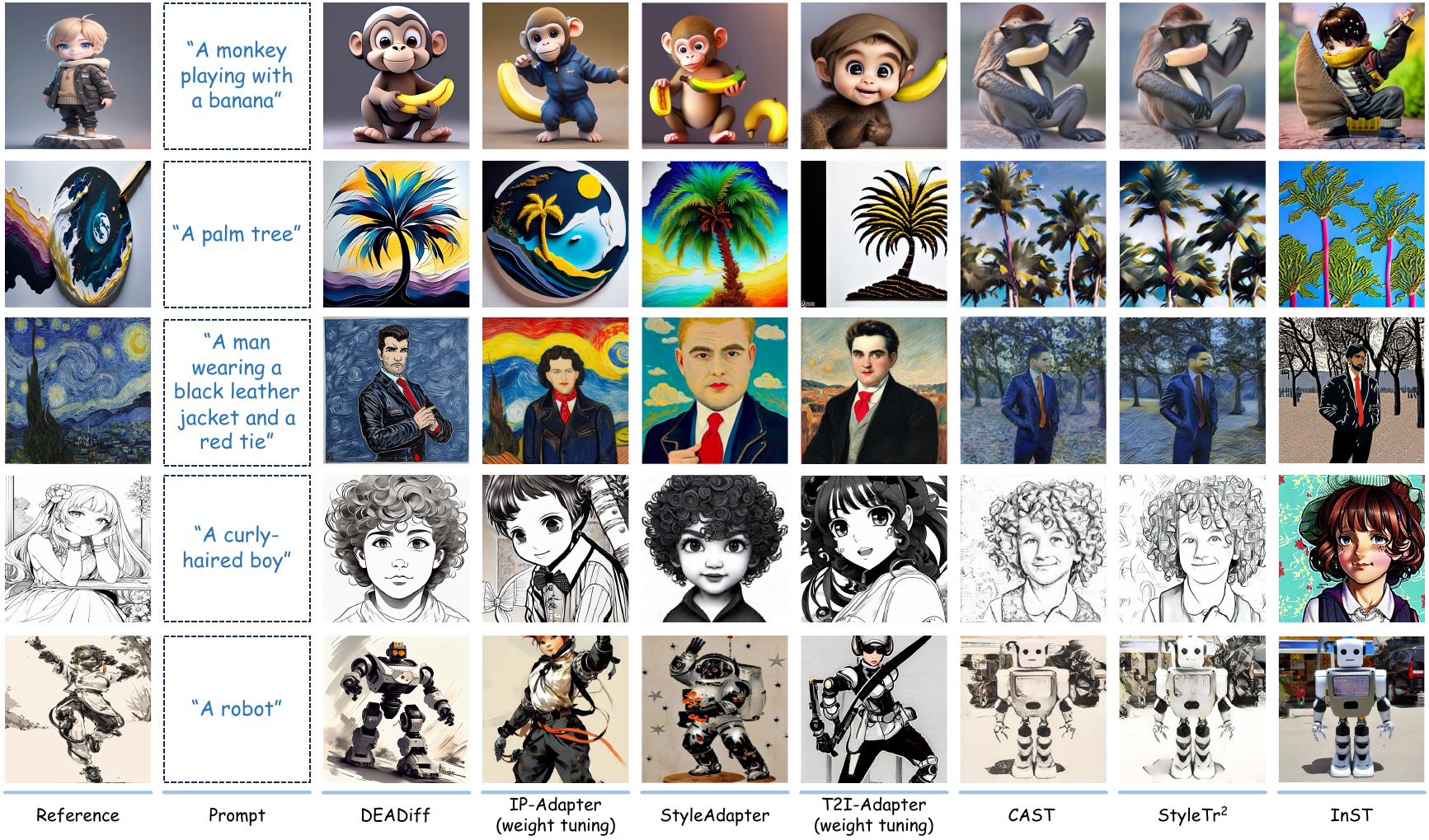}
    \vspace{-7mm}
    \caption{Qualitative comparison with the state-of-the-art methods. Zoom in for better visualization.}
    \label{fig:main-result}
    \vspace{-5mm}
\end{figure*}

\noindent\textbf{Datasets.} We use self-constructed datasets as introduced in~\cref{sec:3.3} to train our model.
The initial dataset with 1.06 million image-text pairs is prepared for the reconstruction task.
The style representation learning task is trained using 160000 pairs of images with the same style, while the semantic representation learning task is trained using 1.06 million pairs of images with the same semantics.
Please refer to the supplementary material for more detailed information about self-constructed datasets.
To evaluate the effectiveness of DEADiff, we construct an evaluation set comprising 32 style images collected from the WikiArt dataset~\cite{tan2018improved} and the Civitai platform.
We exclude text prompts with redundant subjects released in StyleAdapter~\cite{wang2023styleadapter}, slimming down the original 52 to a final 35.
We follow the practice of StyleAdapter, employing Stable Diffusion v1.5 to generate content images corresponding to these 35 text prompts, facilitating comparison with style transfer methods, such as CAST~\cite{zhang2022domain} and StyleTR$^2$~\cite{deng2022stytr2}.

\noindent\textbf{Evaluation Metrics.}
In the absence of a precise and suitable metric for assessing style similarity (SS), we propose a more reasonable approach as elaborated in~\cref{sec:qc}.
Additionally, we determine the cosine similarity within the CLIP text-image embedding space between the textual prompts and their corresponding synthesized images, indicative of the text alignment capability (TA).
We also report the results for the image quality (IQ) of each method.
Finally, to eliminate the interference caused by randomness in the objective metric calculation, we conduct a user study to reflect the subjective preference (SP) for the results.

\subsection{Comparison with State-of-the-Arts}

In this section, we compare our method with the state-of-the-art methods, including optimization-free approaches such as CAST\cite{zhang2022domain}, StyleTr$^2$\cite{deng2022stytr2}, T2I-Adapter\cite{mou2023t2i}, IP-Adapter\cite{ye2023ip} and StyleAdapter\cite{wang2023styleadapter}, as well as optimization-based methods like InST\cite{zhang2023inversion}. It should be noted that since StyleAdapter is not open-sourced, we directly use the results from its released paper for demonstration.

\noindent\textbf{Qualitative Comparisons.} \cref{fig:main-result} illustrates the comparison results with the state-of-the-art methods.
From this figure, we can discern several noteworthy observations.
Firstly, the content image-based style transfer methods, such as CAST~\cite{zhang2022domain} and StyleTr$^2$~\cite{deng2022stytr2}, which do not leverage diffusion models, bypass the issue of reduced text control.
However, they merely execute the straightforward color transfer and refrain from engaging more distinctive features like brush strokes and textures from the reference image, leading to noticeable artifacts in each synthesized outcome.
Consequently, when such methods encounter scenarios with intricate style references and sizable complexity in content image structures, their style transfer ability notably diminishes.
Additionally, for methods trained with the reconstruction objective utilizing diffusion models, whether they are optimization-based (InST~\cite{zhang2023inversion}) or optimization-free (T2I-Adapter~\cite{mou2023t2i}), they generally face semantics interference from the style images in the generated results, as shown in the first and fourth rows of~\cref{fig:main-result}.
This aligns with our previous analysis of the semantics conflict issue.
Thirdly, while the subsequent improved work, StyleAdapter~\cite{wang2023styleadapter}, effectively tackles the problem of semantics conflicts, the style it learns is suboptimal.
It loses the detailed strokes and textures of the reference, and there are also noticeable differences in color.
Lastly, IP-Adapter~\cite{ye2023ip} with meticulous weight tuning for each reference image can achieve decent results, but its synthesized outputs either introduce some semantics from the reference images or suffer from style degradation.
On the contrary, our method not only better adheres to the textual prompts but also significantly preserves the overall style and detailed textures of the reference image, with very minor differences in the color tones.

\begin{table}
  \centering
  \begin{tabular}{@{}lcccc@{}}
    \toprule
    Method & SS$\uparrow$ & IQ$\uparrow$ & TA$\uparrow$ & SP$\uparrow$ \\
    \midrule
    InST~\cite{zhang2023inversion} & 0.215 & 5.148 & 0.237 & 6.3 \\
    CAST~\cite{zhang2022domain} & 0.224 & 4.922 & \underline{0.282} & 8.7 \\
    StyTr$^2$~\cite{deng2022stytr2} & 0.214 & 5.037 & \underline{0.282} & \underline{13.1} \\
    T2I-Adapter~\cite{mou2023t2i} & 0.241 & 5.500 & 0.224 & 2.7 \\
    IP-Adapter~\cite{ye2023ip} & \textbf{0.274} & \underline{5.598} & 0.155 & - \\
    DEADiff & \underline{0.229} & \textbf{5.840} & \textbf{0.284} & \textbf{69.0} \\
    \bottomrule
  \end{tabular}
  \vspace{-2mm}
  \caption{Quantitative comparison with the state-of-the-arts.}
  \label{tab:main}
  \vspace{-3mm}
\end{table}

\begin{figure}
    \centering
    \includegraphics[width=1.0\linewidth]{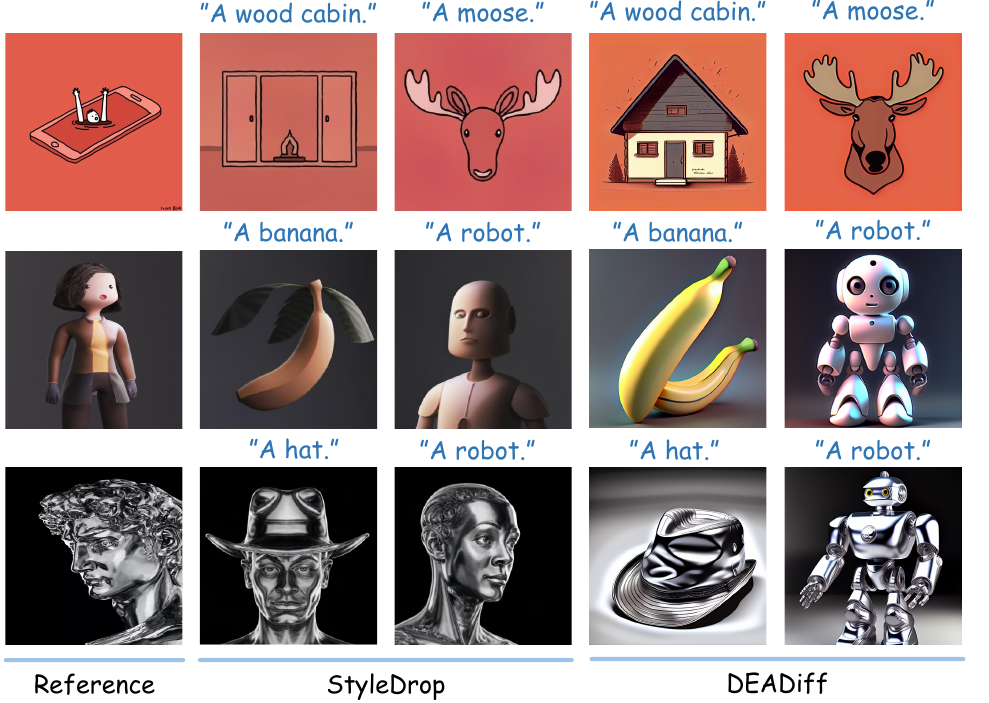}
    \vspace{-6mm}
    \caption{Visual comparison between StyleDrop and \textit{DEADiff}.}
    \label{fig:styledrop}
    \vspace{-5mm}
\end{figure}

\noindent\textbf{Quantitative Comparisons.} \cref{tab:main} presents the style similarity, image quality, text alignment and the overall subjective preference of our method compared with the state-of-the-art methods on the evaluation set we constructed.
We draw several conclusions from this table.
First, aside from T2I-Adapter~\cite{mou2023t2i} and IP-Adapter~\cite{ye2023ip} without meticulous weight tuning (whose generated results are often a reorganization of the reference images, as evidenced by their low text alignment scores), we achieve the highest style similarity, demonstrating that our method indeed effectively captures the overall style of the reference images to some extent.
Second, our method achieves comparable text alignment to the two SD-based methods for generating content images, CAST~\cite{zhang2022domain} and StyTr$^2$~\cite{deng2022stytr2}.
This indicates that our method does not compromise the original text control capabilities of SD while learning the style of the reference images.
Third, the substantial advantage reflected
in the image quality metric compared to all other methods corroborates the practicality of our approach.
Furthermore, as shown in the rightmost column of~\cref{tab:main}, users demonstrate a significantly greater preference for our method over all other ones.
More detailed results and explanations could be found in supplement materials~\cref{sec:qc} and~\cref{sec:us}.
In summary, \textit{DEADiff} achieves an optimal balance between text fidelity and image similarity with the most pleasing image quality.

\noindent\textbf{Comparison with StyleDrop~\cite{sohn2023styledrop}}
Additionally, \cref{fig:styledrop} presents a visual comparison between our method and StyleDrop.
Overall, although \textit{DEADiff} is slightly inferior to optimization-based StyleDrop in terms of color accuracy, it achieves comparable or even better results in terms of artistic style and fidelity to the text.
The cabin, hat, and robot generated by \textit{DEADiff} are more appropriate and do not suffer from semantic interference inherently present in the reference image.
This demonstrates the critical role of disentangling semantics from the reference image.

\begin{figure}
    \centering
    \includegraphics[width=1.0\linewidth]{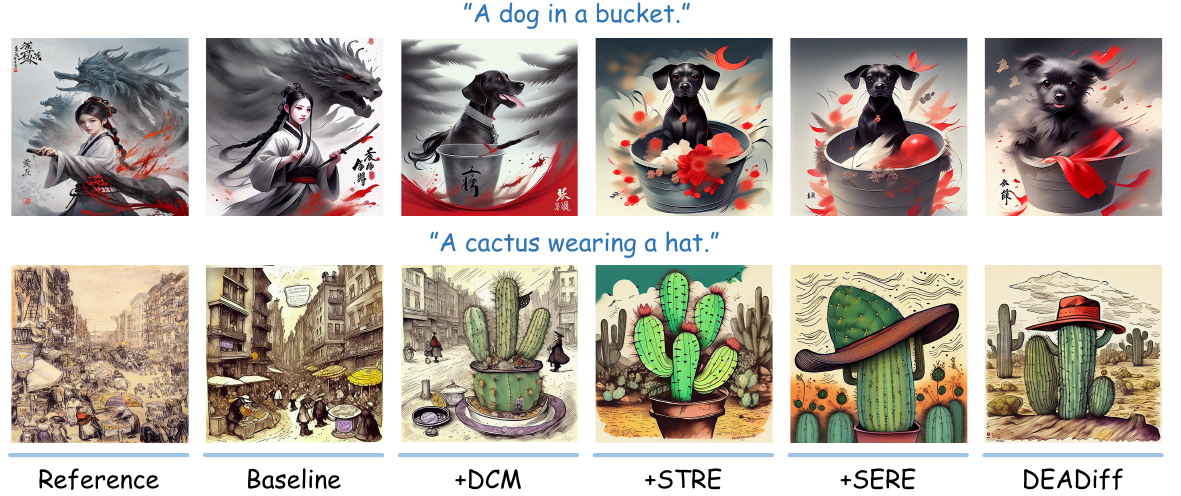}
    \vspace{-5mm}
    \caption{Representative visual results under all configurations listed in~\cref{tab:ablation}.}
    \label{fig:ablation}
\end{figure}

\begin{table}
  \centering
  \begin{tabular}{@{}lcc@{}}
    \toprule
    Method & Style Similarity$\uparrow$ & Text Alignment$\uparrow$ \\
    \midrule
    Baseline & \textbf{0.274} & 0.148 \\
    + DCM & 0.259 & 0.224 \\
    + STRE & 0.222 & 0.286 \\
    + SERE & 0.221 & 0.287 \\
    DEADiff & 0.224 & \textbf{0.289} \\
    \bottomrule
  \end{tabular}
  \vspace{-2mm}
  \caption{Quantitative results from gradually increasing components with \textit{DEADiff}.}
  \label{tab:ablation}
  \vspace{-5mm}
\end{table}

\begin{figure}
    \centering
    \includegraphics[width=0.75\linewidth]{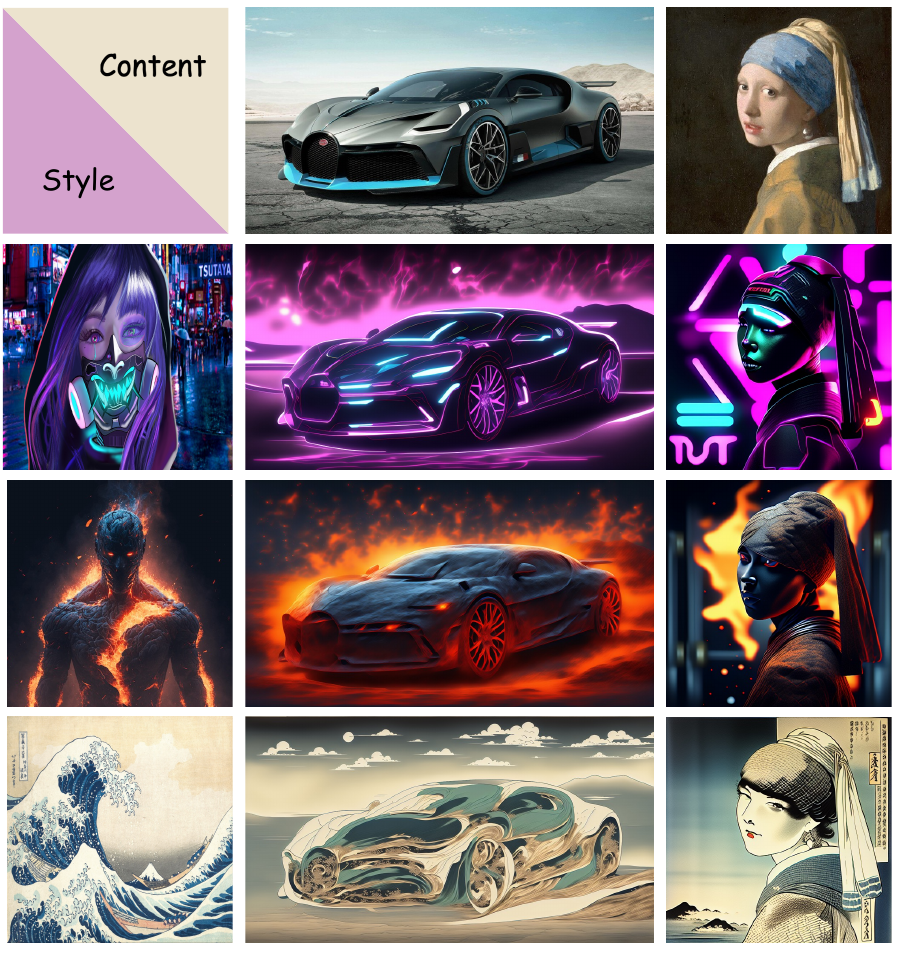}
    \vspace{-2mm}
    \caption{Visual results for content image-based stylization.}
    \label{fig:style_trasfer}
    \vspace{-5mm}
\end{figure}

\begin{figure*}
    \centering
    \includegraphics[width=1.0\linewidth]{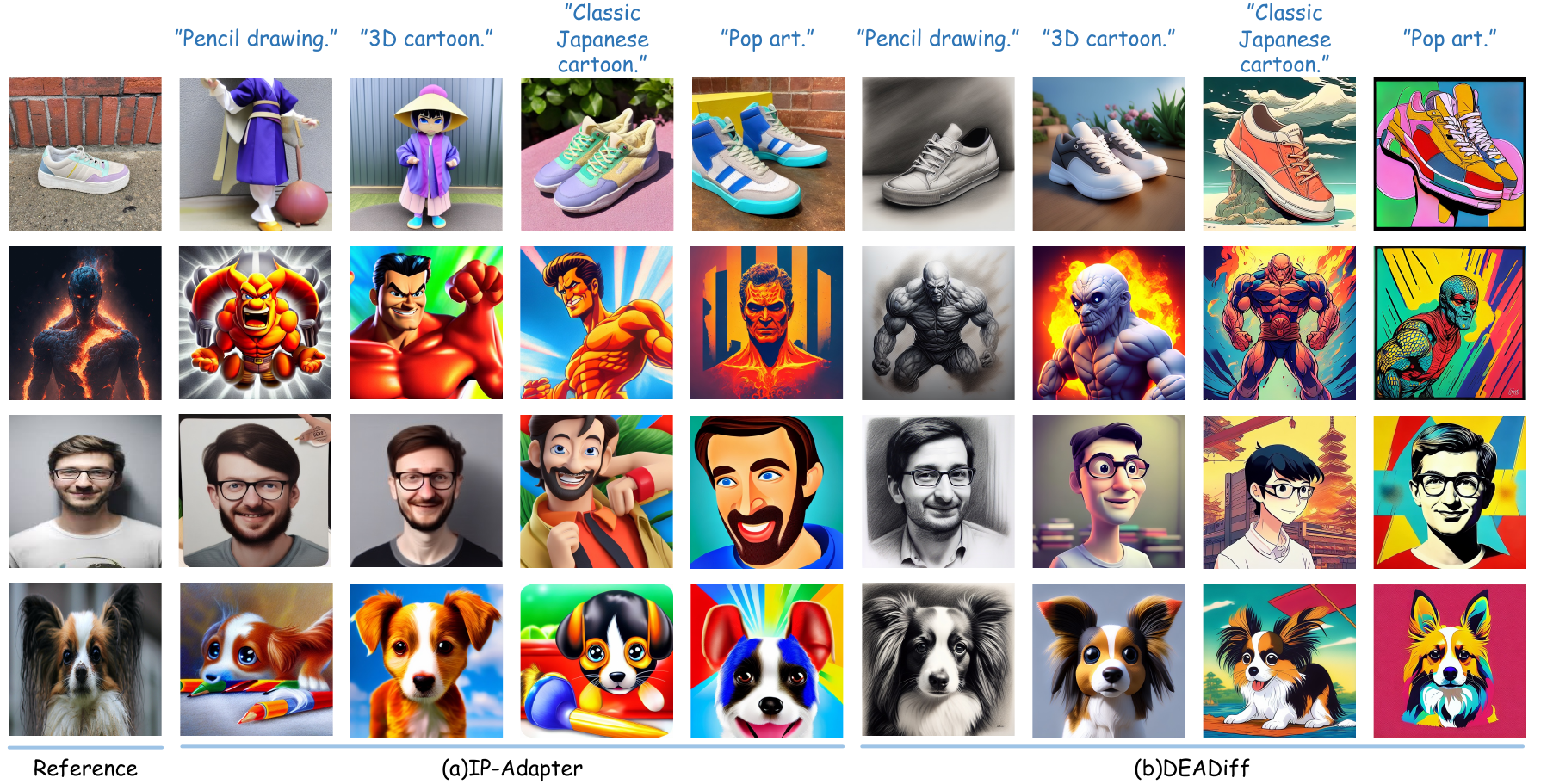}
    \vspace{-6mm}
    \caption{Visual results for the stylization of reference semantics. Note that we reduce the weight of the image condition in IP-Adapter~\cite{ye2023ip} to enhance the efficacy of text prompts in controlling style.}
    \label{fig:semantic_stylization}
    \vspace{-3mm}
\end{figure*}

\begin{figure}
    \centering
    \includegraphics[width=1.0\linewidth]{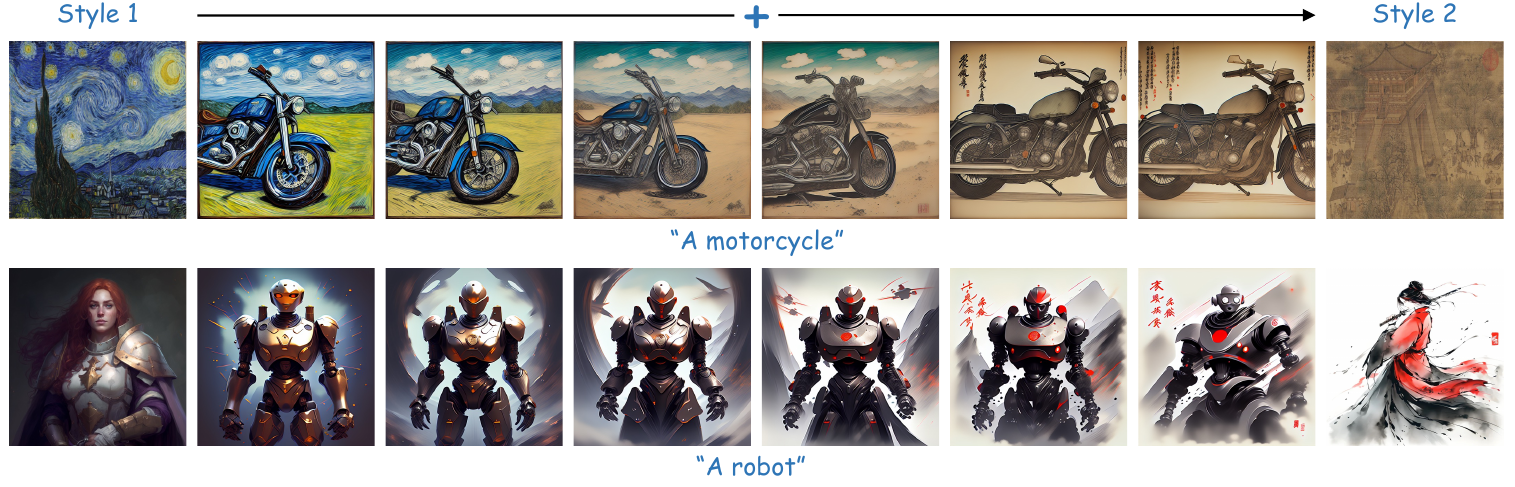}
    \vspace{-6mm}
    \caption{Visual results for style mixing.}
    \label{fig:style_mixing}
\end{figure}

\begin{figure}
    \centering    \includegraphics[width=0.8\linewidth]{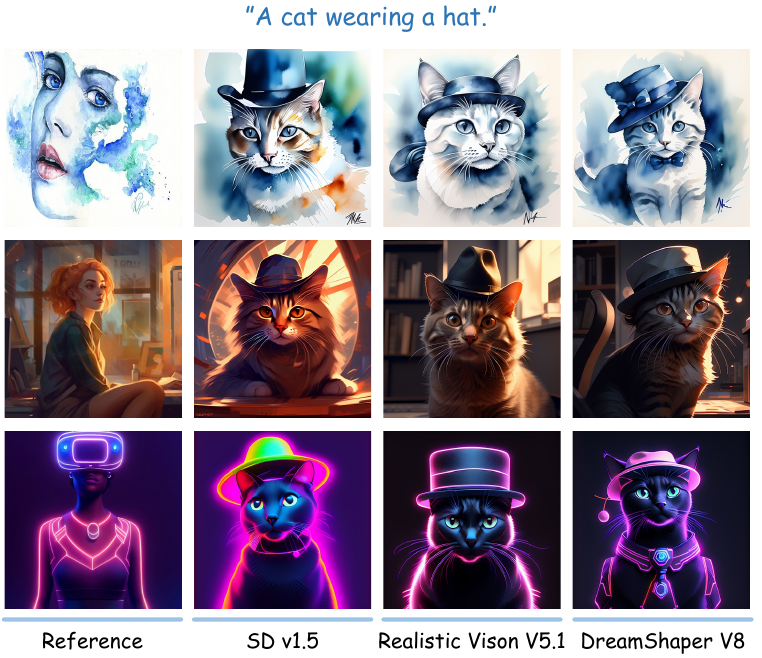}
    \vspace{-2mm}
    \caption{Visual results for substituting the denoising U-Net.}
    \label{fig:base_model}
    \vspace{-3mm}
\end{figure}

\subsection{Ablation Study}

To comprehend the roles each component plays within \textit{DEADiff}, we conduct a series of ablation studies.
\cref{tab:ablation} presents the quantitative results under all configurations, whereas~\cref{fig:ablation} enumerates representative visual outcomes.
Note that the baseline refers to injecting image features extracted by Q-Former into all cross-attention layers of the U-Net~\cite{ronneberger2015u}, which is trained with the reconstruction paradigm.
Each configuration is assessed on the evaluation set after training 50,000 iterations.

\noindent\textbf{Disentangled Conditioning Mechanism.} Combining the top two rows of~\cref{tab:ablation} and the second and third columns of~\cref{fig:ablation}, it is clear that the reconstruction training paradigm inevitably introduces semantics from the reference image, masking the control capabilities of text prompts.
Even though DCM does enhance it by capitalizing on U-Net's characteristic of responding differently to conditions at different layers, as evidenced by the visual results and higher text alignment, the semantic component from image features still conflicts with text semantics.

\noindent\textbf{Dual Decoupling Representation Extraction.} Referring to the bottom three rows of~\cref{tab:ablation} and the rightmost three columns of~\cref{fig:ablation}, we observe a notable enhancement in text editability compared to the former DCM and further progressive improvement.
Specifically, STRE (the third row in~\cref{tab:ablation}) introduces a non-reconstructive training paradigm, allowing the features extracted by Q-Former to focus more on the style information of the reference image, thereby reducing the semantic components contained within.
Hence, the content of the reference image immediately disappears from the generated results, as depicted in the fourth column in~\cref{fig:ablation}.
In addition, while the introduction of SERE (the penultimate row in~\cref{tab:ablation}) seems to have limited impact on the results, its combination with STRE (the last row in~\cref{tab:ablation}) to reconstruct the original image ensures that the extracted two representations are decoupled, complementing each other without omissions.
As shown by the last column in~\cref{fig:ablation}, the text control capabilities are perfectly manifested while fully replicating the style of the reference image with the overall \textit{DEADiff}.

\subsection{Applications}

\noindent\textbf{Combination with ControlNet~\cite{zhang2023adding}.} \textit{DEADiff} supports all types of ControlNets native to SD v1.5. Taking depth ControlNet as an example, \cref{fig:style_trasfer} demonstrates the impressive effects of stylization while maintaining the layout.
\textit{DEADiff} has a wide application scope. In this section, we enumerate a few of its typical applications.

\noindent\textbf{Stylization of reference semantics.} Since \textit{DEADiff} can extract the semantic representation of the reference image, it can stylize the semantic objects in the reference image through text prompts.
As shown in~\cref{fig:semantic_stylization}, the stylization effects are significantly superior to that of IP-Adapter~\cite{ye2023ip}.

\noindent\textbf{Style mixing.} \textit{DEADiff} is capable of blending styles from multiple reference images.
\cref{fig:style_mixing} shows its progressive changing effects under the different control exerted by two reference images.

\noindent\textbf{Switch of the base T2I model.} Since \textit{DEADiff} does not optimize the base T2I models, it can directly switch between different base models to generate different stylization results, as shown in~\cref{fig:base_model}.
\section{Conclusion}

In this paper, we delve into the reasons for the decline in text control capabilities of existing encoder-based stylized diffusion models and subsequently propose the targeted design of \textit{DEADiff}. It includes a dual decoupling representation extraction mechanism and a disentangled conditioning mechanism. Empirical evidence demonstrates that \textit{DEADiff} is capable of attaining an optimal equilibrium between stylization capabilities and text control. Future work could aim to further enhance style similarity and decouple instance-level semantic information.

{
    \small
    \bibliographystyle{ieeenat_fullname}
    \bibliography{main}
}
\clearpage
\setcounter{page}{1}
\section{Supplementary}



\subsection{Quantitative Comparisons}
\label{sec:qc}
Given that \textit{DEADiff} is proposed specifically to address the issue of text controllability loss inherent in encoder-based methods, we primarily emphasize the quantitative metric of text alignment in the main paper.
Below, we additionally provide a quantitative comparison of the style similarity and image quality between \textit{DEADiff} and the state-of-the-art methods, as illustrated by~\cref{tab:qc}.

\noindent\textbf{Evaluation Metrics.}

\noindent\textbf{Style Similarity:} We propose a more reasonable approach to measure style similarity.
Specifically, the procedure begins with using the CLIP Interrogator~\footnote{\href{https://github.com/pharmapsychotic/clip-interrogator}{https://github.com/pharmapsychotic/clip-interrogator}} to generate the optimal text prompts that align with the reference image.
Subsequently, we filter out the prompts related to the content of the reference image and compute the cosine similarity between the remaining prompts and the generated image within the CLIP text-image embedding space.
The computational result denotes the style similarity, effectively mitigating interference from the content of the reference image.

\noindent\textbf{Image Quality:} We adopt a prediction model named LAION-Aesthetics Predictor V2~\footnote{\href{https://github.com/christophschuhmann/improved-aesthetic-predictor}{https://github.com/christophschuhmann/improved-aesthetic-predictor}} to assess the quality of images generated by each method.

\noindent\textbf{Text Alignment:} We determine the cosine similarity within the CLIP text-image embedding space between the textual prompts and their corresponding synthesized images, indicative of the text alignment capability.



\begin{figure}
    \centering
    \includegraphics[width=1.0\linewidth]{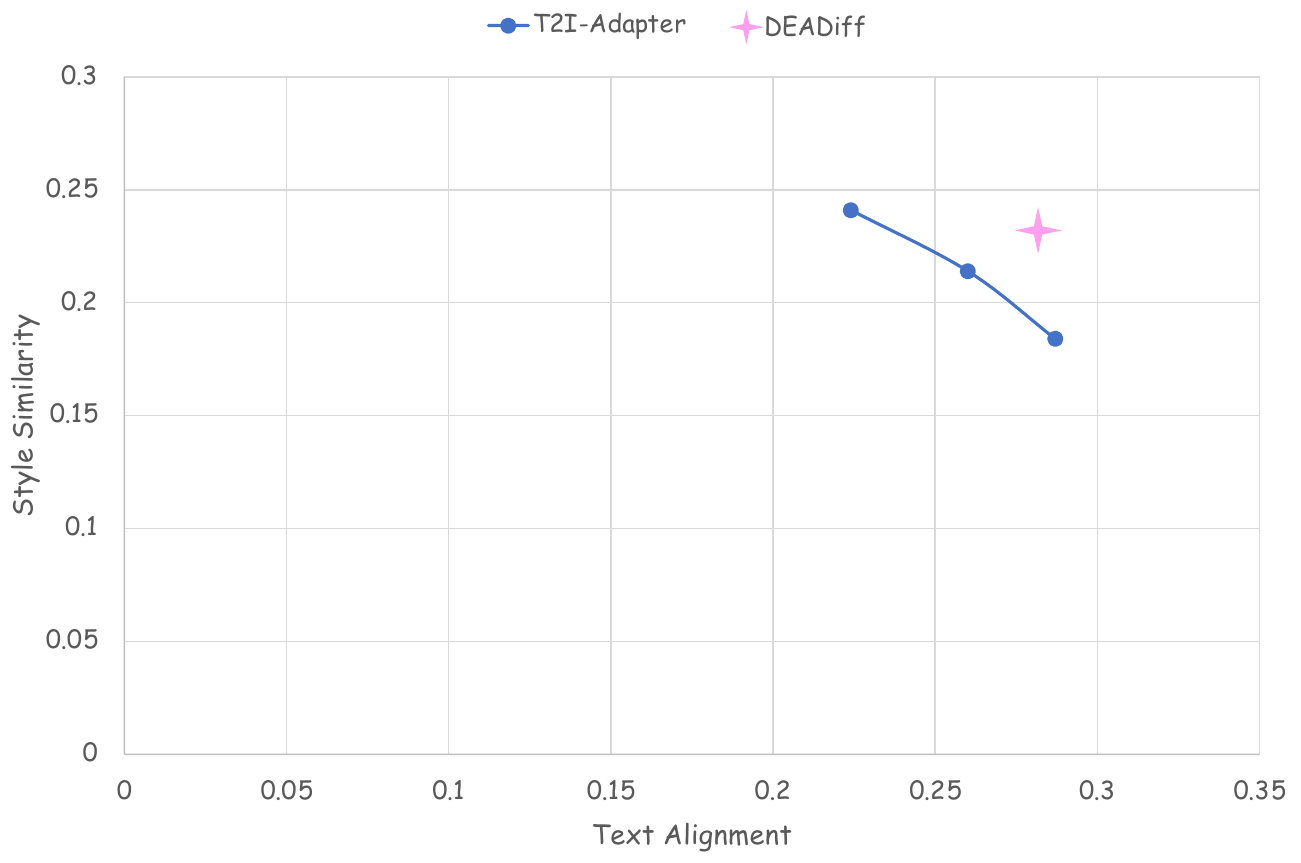}
    \caption{Quantitative comparison between \textit{DEADiff} and the trade-off curve of T2I-Adapter.}
    \label{fig:trade-off}
\end{figure}

Differing from~\cref{tab:main}, we not only list the quantitative results of T2I-Adapter~\cite{mou2023t2i} at the default image condition weight of 1.0, but also provide the results when the image condition weight is set to 0.9 and 0.8 in~\cref{tab:qc}.
Evidently, T2I-Adapter, under different image condition weights, exhibits a clear trade-off between style similarity and text alignment.
When the image condition weight is overly large, \textit{e.g.}, 1.0, the generated image essentially becomes a reorganization of the reference image.
This leads to a high style similarity (0.241) but significantly weakens text controllability (0.224), as introduced in~\cref{sec:intro}.
However, if we reduce this weight, the style similarity will drop rapidly to 0.184.
\cref{fig:trade-off} provides an intuitive illustration that \textit{DEADiff} is situated outside T2I-Adapter's trade-off curve, thereby demonstrating its enhanced ability to strike a balance between style similarity and text control capability.
Moreover, \textit{DEADiff} outperforms other top-performing methods in both style similarity and text alignment, including CAST, StyTr$^2$ and InST, further confirming the effectiveness of our approach.
Meanwhile, the substantial advantage reflected in the image quality metric compared to all other methods corroborates the practicality of our approach.

\begin{table*}[!t]
  \centering
  \begin{tabular}{@{}lccc@{}}
    \toprule
    Method & Style Similarity$\uparrow$ & Image Quality$\uparrow$ & Text Alignment$\uparrow$ \\
    \midrule
    InST~\cite{zhang2023inversion} & 0.215 & 5.148 & 0.237 \\
    CAST~\cite{zhang2022domain} & 0.224 & 4.922 & 0.282 \\
    StyTr$^2$~\cite{deng2022stytr2} & 0.214 & 5.037 & 0.282 \\
    T2I-Adapter 1.0~\cite{mou2023t2i} & \textbf{0.241} & 5.500 & 0.224 \\
    T2I-Adapter 0.9~\cite{mou2023t2i} & 0.214 & 5.534 & 0.260 \\
    T2I-Adapter 0.8~\cite{mou2023t2i} & 0.184 & \underline{5.580} & \textbf{0.287} \\
    DEADiff & \underline{0.229} & \textbf{5.840} & \underline{0.284} \\
    \bottomrule
  \end{tabular}
  \caption{Quantitative comparison of style similarity,  image quality and text alignment with the state-of-the-art methods. \textbf{Bold} numbers denote the best results, while the \underline{underlined} numbers denote the second best results. We show different results for T2I-Adapter with three varying condition weights: 1.0, 0.9, and 0.8, which presents an obvious trade-off between style similarity and text alignment.}
  \label{tab:qc}
\end{table*}

\begin{table*}
  \centering
  \begin{tabular}{@{}lcccc@{}}
    \toprule
    Aspect & Style Similarity$\uparrow$ & Image Quality$\uparrow$ & Text Alignment$\uparrow$ & Overall$\uparrow$ \\
    \midrule
    InST~\cite{zhang2023inversion} & 7.8 & 8.5 & 11.9 & 6.3 \\
    CAST~\cite{zhang2022domain} & 8.7 & 9.3 & 10.5 & 8.7 \\
    StyTr$^2$~\cite{deng2022stytr2} & 16.1 & 11.5 & 13.9 & 13.1 \\
    T2I-Adapter~\cite{mou2023t2i} & 1.9 & 8.1 & 7.5 & 2.7 \\
    DEADiff & \textbf{65.4} & \textbf{62.5} & \textbf{56.2} & \textbf{69.0} \\
    \bottomrule
  \end{tabular}
  \caption{Results for the user study in percentages.}
  \label{tab:us}
\end{table*}

\subsection{User Study}
\label{sec:us}
In addition to objective evaluations, we have also designed a user study to subjectively assess the practical performance of various methods.
Given 18 style reference images from Civitai~\footnote{\href{https://civitai.com}{https://civitai.com}}, we employed CAST~\cite{zhang2022domain}, InST~\cite{zhang2023inversion}, StyTr$^2$~\cite{deng2022stytr2}, T2I-Adapter~\cite{mou2023t2i}, and \textit{DEADiff} to separately generate corresponding stylized results.
Specifically, we utilized a total of 21 distinct text prompts.
Thus, apart from three reference images corresponding to two prompts each, the remaining 15 reference images and 15 prompts are directly matched one-to-one.
We asked 24 users from diverse backgrounds to evaluate the generated results in terms of text-image alignment, image quality, and style similarity, and to provide their overall preference considering these three aspects.
Consequently, we have obtained a total of 2016 voting results.
The final results are displayed in~\cref{tab:us}.
\textit{DEADiff} outperforms all state-of-the-art methods on three evaluation aspects and the overall preference with a big margin, which demonstrates the broad application prospects of our method.

\begin{figure}
    \centering
    \includegraphics[width=1.0\linewidth]{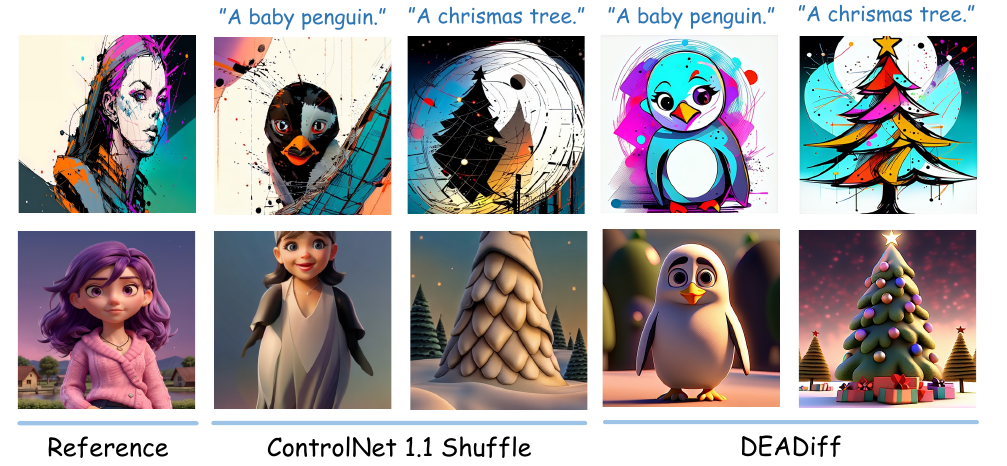}
    \caption{Visual comparison between ControlNet 1.1 Shuffle and \textit{DEADiff}.}
    \label{fig:shuffle}
    \vspace{-5mm}
\end{figure}

\subsection{Inference Efficiency}
Despite \textit{DEADiff} adding 1900 MB to the memory occupation, the increase in average inference time on one A100-80G GPU is only marginal, as shown in~\cref{tab:must}.

\subsection{Comparison with ControlNet 1.1 Shuffle}
We compare our method with ControlNet 1.1 Shuffle~\footnote{\href{https://github.com/lllyasviel/ControlNet-v1-1-nightly\#controlnet-11-shuffle}{https://github.com/lllyasviel/ControlNet-v1-1-nightly\#controlnet-11-shuffle}} and present the results in~\cref{fig:shuffle}.
It is clear that our method outperforms ControlNet 1.1 Shuffle in carving the style of the reference image, fidelity to the text, and generated image quality.

\begin{table}
  \centering
  \resizebox{\columnwidth}{!}{
  \begin{tabular}{@{}lccc@{}}
    \toprule
    Model & SD & ControlNet 1.1 Shuffle & \textit{DEADiff} \\
    \midrule
    Memory (MB) & 7774 & 10986 & 9674 \\
    50-Step DDIM Time on A100 (s) & 2.28 & 3.00 & 2.43 \\
    \bottomrule
  \end{tabular}}
  \caption{Memory usage and sampling time on 1 A100-80G GPU.}
  \label{tab:must}
\end{table}

\subsection{Combination with DreamBooth/LoRA}
\begin{figure}
    \centering
    \includegraphics[width=1.0\linewidth]{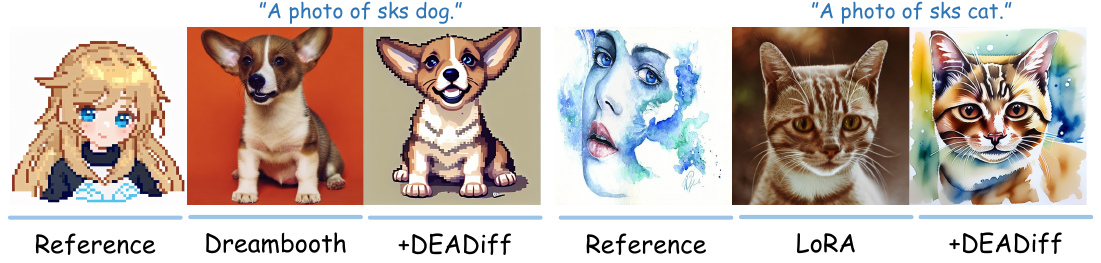}
    \vspace{-3mm}
    \captionof{figure}{Stylize the Dreambooth/LoRA customized subject.}
    \label{fig:dreambooth_lora}
\end{figure}

As the original U-Net parameters are frozen, our method is well compatible with DreamBooth\&LoRA for extension. \cref{fig:dreambooth_lora} shows an example of using DreamBooth/LoRA to control the subject (the dog and the cat) and DEADiff to control the style.

\subsection{More Examples}
\begin{figure*}
    \centering
    \includegraphics[width=0.97\linewidth]{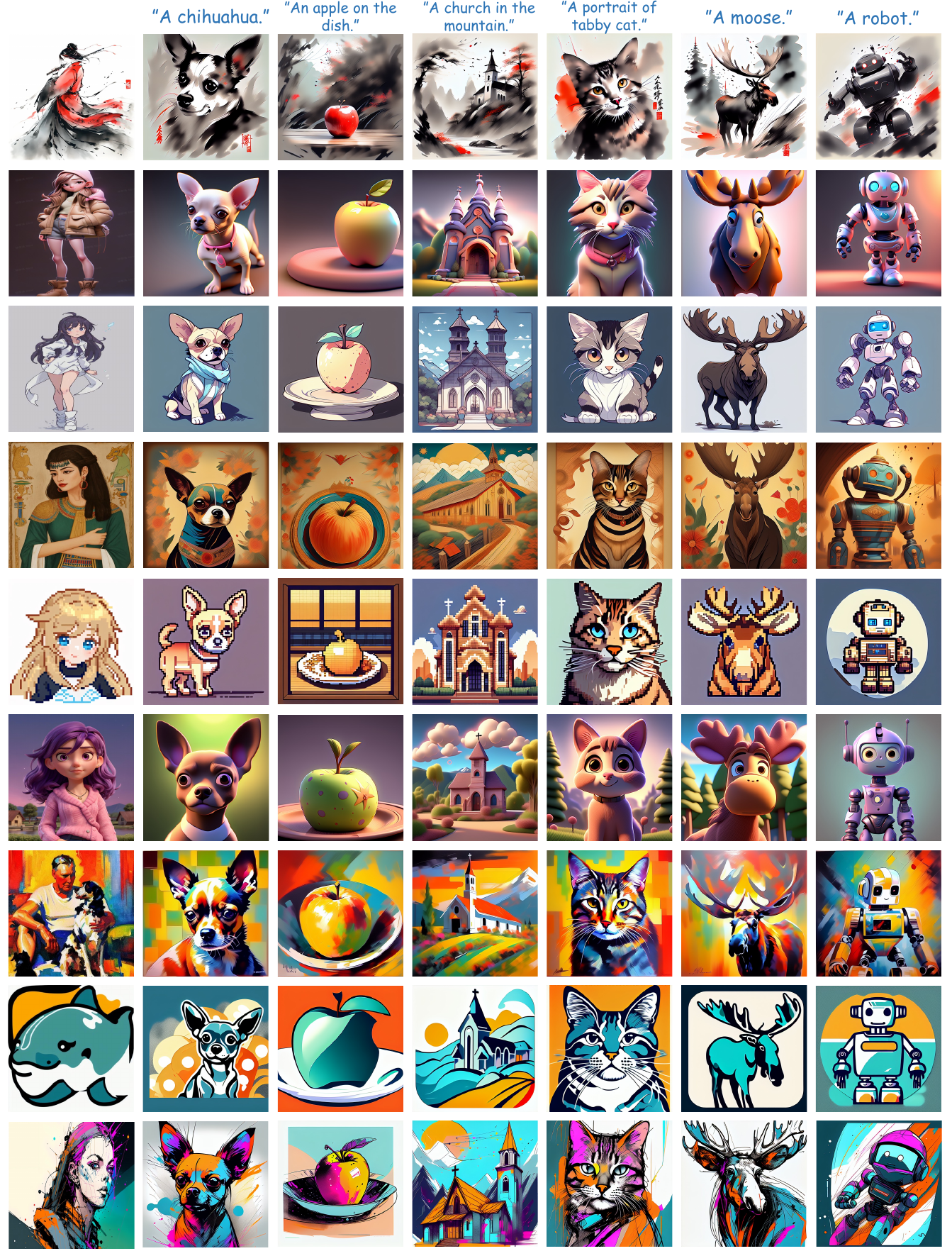}
    \caption{Additional visualization results for \textit{DEADiff}. Our method can synthesize high-quality images that are capable of imitating the reference style and following the instructions of text prompts.}
    \label{fig:more}
    \vspace{-5mm}
\end{figure*}

To show the effectiveness and universality of our method, we present more visualization results in~\cref{fig:more}.

\end{document}